\pdfoutput=1

\documentclass[11pt]{article}

\usepackage{ACL2023}

\usepackage{times}
\usepackage{latexsym}

\usepackage[T1]{fontenc}

\usepackage[utf8]{inputenc}

\usepackage{microtype}

\usepackage{inconsolata}

\usepackage{CJKutf8}

\usepackage{booktabs}
\usepackage{graphicx}
\usepackage{multirow}
\usepackage{amsmath}
\usepackage{amssymb}
\usepackage{bm}

\author{Leonhard Hennig ~~~ Philippe Thomas ~~~ Sebastian Möller\\
German Research Center for Artificial Intelligence (DFKI)  \\
Speech and Language Technology Lab\\
\{\textit{leonhard.hennig, philippe.thomas, sebastian.moeller}\}@\textit{dfki.de}}

\title{MultiTACRED: A Multilingual Version of the TAC Relation Extraction Dataset}

\begin{document}

\maketitle

\begin{abstract}
Relation extraction (RE) is a fundamental task in information extraction, whose extension to multilingual settings has been hindered by the lack of supervised resources comparable in size to large English datasets such as \textsl{TACRED}~\cite{zhang-etal-2017-position}. To address this gap, we introduce the \textsl{MultiTACRED} dataset, covering 12 typologically diverse languages from 9 language families, which is created by machine-translating \textsl{TACRED} instances and automatically projecting their entity annotations. We analyze translation and annotation projection quality, identify error categories, and experimentally evaluate fine-tuned pretrained mono- and multilingual language models in common transfer learning scenarios. Our analyses show that machine translation is a viable strategy to transfer RE instances, with native speakers judging more than 83\% of the translated instances to be linguistically and semantically acceptable. We find monolingual RE model performance to be comparable to the English original for many of the target languages, and that multilingual models trained on a combination of English and target language data can outperform their monolingual counterparts. However, we also observe a variety of translation and annotation projection errors, both due to the MT systems and linguistic features of the target languages, such as pronoun-dropping, compounding and inflection, that degrade dataset quality and RE model performance.
\end{abstract}

\section{Introduction}
\begin{figure*}[ht!]
\centering
\includegraphics[width=0.98\textwidth,keepaspectratio,trim={0.8cm 3.1cm 0.3cm 3.2cm},clip]{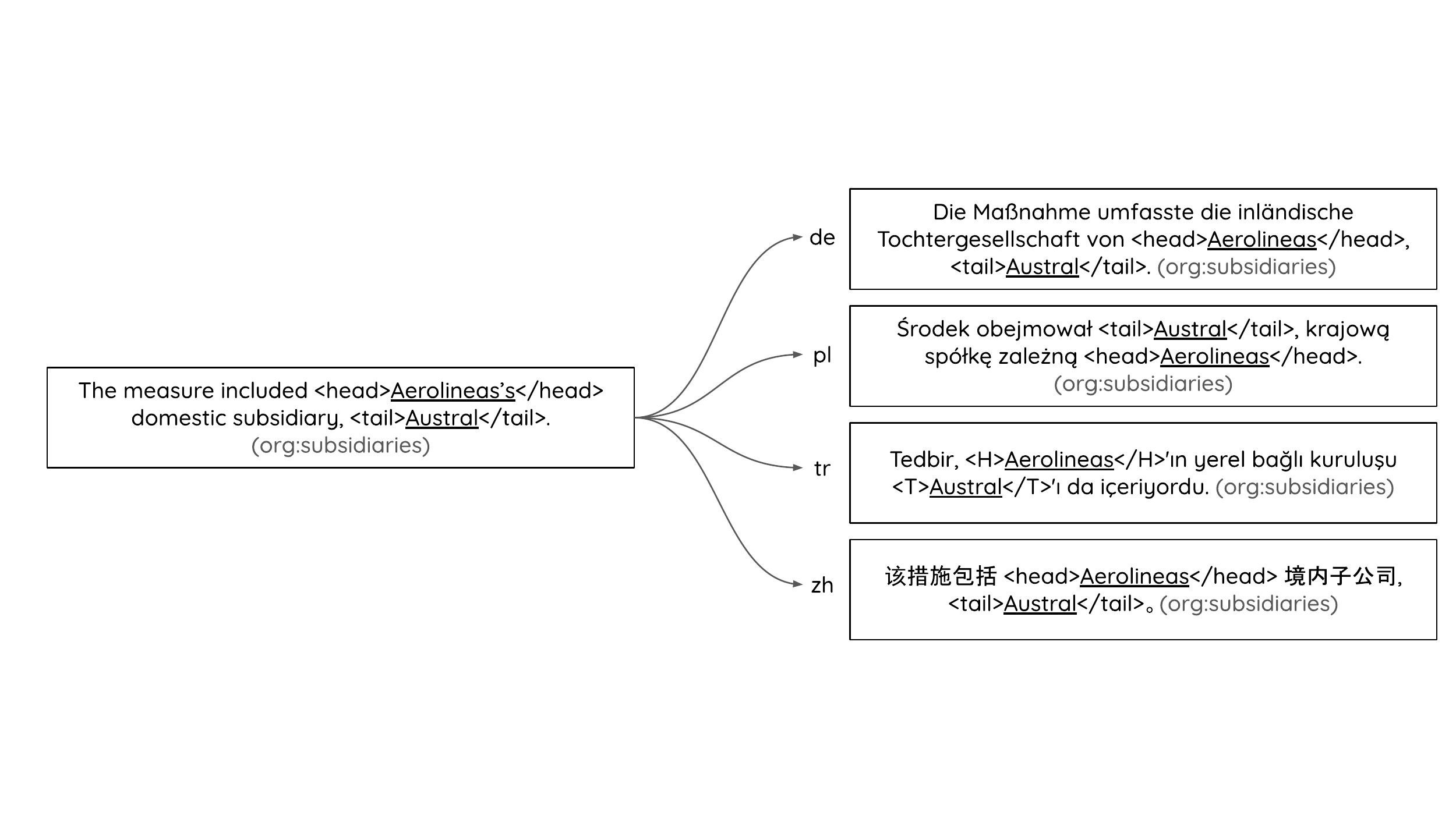}
\caption{Example translations from English to German, Polish, Turkish and Chinese with XML markup for the head and tail entities to project relation argument annotations.}
\label{fig:translation_example}
\end{figure*}

Relation extraction (RE), defined as the task of identifying and classifying semantic relationships between entities from text (cf.\ Figure~\ref{fig:translation_example}), is a fundamental task in information extraction~\cite{doddington-etal-2004-automatic}. Extending RE to multilingual settings has recently received increased interest~\cite{zou-etal-2018-adversarial,nag-etal-2021-data-bootstrap,chen-etal-2022-multilingual}, both to address the urgent need for more inclusive NLP systems that cover more languages than just English~\cite{ruder-etal-2019-survey,pmlr-v119-hu20b}, as well as to investigate language-specific phenomena and challenges relevant to this task. The main bottleneck for multilingual RE is the lack of supervised resources, comparable in size to large English datasets~\cite{riedel_modeling_2010,zhang-etal-2017-position}, as annotation for new languages is very costly. Most of the few existing multilingual RE datasets are distantly supervised \cite{koksal-ozgur-2020-relx,seganti-etal-2021-multilingual,bhartiya-etal-2021-disrex}, and hence suffer from noisy labels that may reduce the prediction quality of models~\cite{riedel_modeling_2010,xie-etal-2021-revisiting}. Available fully-supervised datasets are small, and cover either very few domain-specific relation types~\cite{arviv-etal-2021-relation,khaldi-etal-2022-hows}, or only a small set of languages~\cite{nag-etal-2021-data-bootstrap}. 

To address this gap, and to incentivize research on supervised multilingual RE, we introduce a multilingual version of one of the most prominent supervised RE datasets, \textsl{TACRED}~\cite{zhang-etal-2017-position}. \textsl{MultiTACRED} is created by machine-translating \textsl{TACRED} instances and automatically projecting their entity annotations. Machine translation is a popular approach for generating data in cross-lingual learning~\cite{pmlr-v119-hu20b,nag-etal-2021-data-bootstrap}. Although the quality of machine-translated data may be lower due to translation and alignment errors~\cite{yarmohammadi-etal-2021-everything}, it has been shown to be beneficial for classification and structured prediction tasks~\cite{pmlr-v119-hu20b,ozaki-etal-2021-project,yarmohammadi-etal-2021-everything}. 

The \textsl{MultiTACRED} dataset we present in this work covers 12 languages from 9 language families.\footnote{ \textsl{MultiTACRED} includes the following language families / languages: German (Germanic); Finnish, Hungarian (Uralic); Spanish, French (Romance); Arabic (Semitic);  Hindi (Indo-Iranic); Japanese (Japonic); Polish, Russian (Slavic); Turkish (Turkic); Chinese (Sino-Tibetan).} We select typologically diverse languages which span a large set of linguistic phenomena such as compounding, inflection and pronoun-drop, and for which a monolingual pretrained language model is available. We automatically and manually analyze translation and annotation projection quality in all target languages, both in general terms and with respect to the RE task, and identify typical error categories for alignment and translation that may affect model performance. We find that overall translation quality is judged to be quite good with respect to the RE task, but that e.g.\ pronoun-dropping, coordination and compounding may cause alignment and semantic errors that result in erroneous instances. In addition, we experimentally evaluate fine-tuned pretrained mono- and multilingual language models (PLM) in common training scenarios, using source language (English), target language, or a mixture of both as training data. We also evaluate an English data fine-tuned model on back-translated test instances to estimate the effect of noise introduced by the MT system on model performance. Our results show that in-language training works well, given a suitable PLM. Cross-lingual zero-shot transfer is acceptable for languages well-represented in the multilingual PLM, and combining English and target language data for training considerably improves performance across the board. 

To summarize, our work aims to answer the following research questions: Can we reaffirm the usefulness of MT and cross-lingual annotation projection, in our study for creating large-scale, high quality multilingual datasets for RE? How do pretrained mono- and multilingual encoders compare to each other, in within-language as well as cross-lingual evaluation scenarios? Answers to these questions can provide insights for understanding language-specific challenges in RE, and further research in cross-lingual representation and transfer learning. The contributions of this paper are:
\begin{itemize}
    \item We introduce \textsl{MultiTACRED}, a translation of the widely used, large-scale \textsl{TACRED} dataset into 12 typologically diverse target languages: Arabic, German, Spanish, French, Finnish, Hindi, Hungarian, Japanese, Polish, Russian, Turkish, and Chinese. %
    \item We present an evaluation of monolingual, cross-lingual, and multilingual models to evaluate target language performance for all 12 languages. %
    \item We present insights into the quality of machine translation for RE, analyzing alignment as well as language-specific errors.
\end{itemize}

\section{Translating TACRED}
\label{sec:translating_tacred}
We first briefly introduce the original \textsl{TACRED} dataset, and then describe the language selection and automatic translation process. We wrap up with a description of the analyses we conduct to verify the translation quality.

\subsection{The TACRED dataset}
\label{subsec:tacred}

The \textsl{TAC} \textsl{R}elation \textsl{E}xtraction \textsl{D}ataset\footnote{\url{https://catalog.ldc.upenn.edu/LDC2018T24}, under a LDC license}, introduced by \citet{zhang-etal-2017-position}, is a fully supervised dataset of sentence-level binary relation mentions. It consists of 106k sentences with entity mention pairs collected from the TAC KBP\footnote{\url{https://tac.nist.gov/2017/KBP/index.html}} evaluations 2009--2014, with the years 2009 to 2012 used for training, 2013 for development, and 2014 for testing. 
Each sentence is annotated with a head and a tail entity mention, and labeled with one of 41 person- and organization-oriented relation types, e.g.\ \emph{per:title}, \emph{org:founded}, or the label \emph{no\_relation} for negative instances. About 79.5\% of the examples are labeled as \emph{no\_relation}.\footnote{The first row of Table~\ref{tab:app_multitacred_dataset_stats} in Appendix~\ref{sec:app_translation_details} summarizes key statistics of the dataset.}
All relation labels were obtained by crowdsourcing, using Amazon Mechanical Turk. Recent work by~\citet{alt-etal-2020-tacred} and \citet{stoica_2021_retacred} improved upon the label quality of the crowd annotations by re-annotating large parts of the dataset.

\subsection{Automatic Translation}
\label{subsec:translation}
We translate the complete \textit{train}, \textit{dev} and \textit{test} splits of \textsl{TACRED} into the target languages, and in addition back-translate the \textit{test} split into English to generate machine-translated English test data. Each instance in the original \textsl{TACRED} dataset is a list of tokens, with the head and tail entity arguments of the potential relation specified via token offsets. For translation, we concatenate tokens with whitespace and convert head and tail entity offsets into XML-style markers to denote the arguments' boundaries, as shown in Figure~\ref{fig:translation_example}. We use the commercial services of DeepL\footnote{\url{https://api.deepl.com/v2/translate}} and Google\footnote{\url{https://translation.googleapis.com/language/translate/v3}}, since both offer the functionality to preserve XML tag markup. Since API costs are similar, we use DeepL for most languages, and only switch to Google for languages not supported by DeepL (at the time we were running the MT). We validate the translated text by checking the syntactic correctness of the XML tag markup, and discard translations with invalid tag structure, e.g.\ missing or invalid head or tail tag pairs. 

After translation, we tokenize the translated text using language-specific tokenizers.\footnote{See Appendix~\ref{sec:app_translation_details} for details.} Finally, we store the translated instances in same JSON format as the original \textsl{TACRED} English dataset, with fields for tokens, entity types and offsets, label and instance id. We can then easily apply the label corrections provided by e.g.~\citet{alt-etal-2020-tacred} or \citet{stoica_2021_retacred} to any target language dataset by applying the respective patch files.

We select target languages to cover a wide set of interesting linguistic phenomena, such as compounding (e.g., German), inflection/derivation (e.g., German, Turkish, Russian), pronoun-dropping (e.g., Spanish, Finnish, Polish), and varying degrees of synthesis (e.g., Turkish, Hungarian vs.\ Chinese). We also try to ensure that there is a monolingual pretrained language model available for each language, which is the case for all languages except Hungarian. The final set of languages in \textsl{MultiTACRED} is: German, Finnish, Hungarian, French, Spanish, Arabic, Hindi, Japanese, Chinese, Polish, Russian, and Turkish. Table~\ref{tab:app_multitacred_dataset_stats} in Appendix~\ref{sec:app_translation_details} lists key statistics per language.

\subsection{Translation Quality Analysis}
\label{subsec:translation_quality_analysis}
To verify the overall quality of the machine-translated data, we  also manually inspect translations. For each language, we randomly sample 100 instances from the \textit{train} split. For each sample instance, we display the source (English) text with entity markup (see Figure~\ref{fig:translation_example} for the format), the target language text with entity markup, and the relation label. 

We then ask native speakers to judge the translations by answering two questions: 
(Q1) Does the translated text meaningfully preserve the semantic relation of the English original, regardless of minor translation errors?\footnote{If necessary, human judges are first introduced to the task of relation extraction. They are also given the list of relations and their official definitions for reference.} (Q2) Is the overall translation linguistically acceptable for a native speaker? Human judges are instructed to read both the English source and the translation carefully, and then to answer the two questions with either \textit{yes} or \textit{no}. They may also add free-text comments, e.g.\ to explain their judgements or to describe translation errors. The samples of each language are judged by a single native speaker. Appendix~\ref{sec:app_translation_analysis} gives additional details. 

In addition, we conduct a manual analysis of the automatically discarded translations, using a similar-sized random sample from the German, Russian and Turkish \emph{train} splits, to identify possible reasons and error categories. These analyses are performed by a single trained linguist per language, who is also a native speaker of that language, with joint discussions to synthesize observations. Results of both analyses are presented in Section~\ref{subsec:results_translation_quality}.

\section{Experiments}
\label{sec:experiments}
In this section, we describe the experiments we conduct to answer the research questions ``How does the performance of language-specific models compare to the English original?'' and ``How does the performance of language-specific models compare to multilingual models such as mBERT trained on the English source data? How does the performance change when including target-language data for training''. We first introduce the training scenarios, and then give details on choice of models and hyperparameters, as well as the training process.

\subsection{Training scenarios}
\label{subsec:training_scenarios}
We evaluate the usefulness of the translated datasets by following the most prevalent approach of framing RE as a sentence-level supervised multi-class classification task. Formally, given a relation set $\mathcal R$ and a text $\bm x=[x_1, x_2, \ldots, x_{n}]$ (where $x_1, \cdots, x_{n}$ are tokens) with two disjoint spans $\bm e_h=[x_i, \ldots, x_j]$ and $\bm e_t=[x_k, \ldots, x_l]$ denoting the head and tail entity mentions, RE aims to predict the relation $r\in\mathcal R$ between $\bm e_h$ and $\bm e_t$, or assign the \textit{no\_relation} class if no relation in $\mathcal R$ holds. Similar to prior work (e.g.,~\citet{nag-etal-2021-data-bootstrap}), we evaluate relation extraction models in several different transfer learning setups, which are described next.

\noindent\textbf{Monolingual} We evaluate the performance of language-specific PLMs for each of the 12 target languages, plus English, where the PLM is supervisedly fine-tuned on the \textit{train} split of the respective language.

\noindent\textbf{Cross-lingual} We evaluate the performance of a multilingual mBERT model on the test split of each of the 12 target languages, plus English, after training on the English \textit{train} split. 

\noindent\textbf{Mixed / Multilingual} We evaluate the performance of a multilingual mBERT model on the test split of each of the 12 target languages, after training on the complete English \textit{train} split and a variable portion of the \textit{train} split of the target language, as suggested e.g.\ by~\citet{nag-etal-2021-data-bootstrap}. We vary the amount of target language data in \{5\%,10\%,20\%,30\%,40\%,50\%,100\%\} of the available  training data. When using 100\%, we are effectively doubling the size of the training set, and ``duplicating'' each training instance. 

\noindent\textbf{Back-translation} Finally, we also evaluate the performance of a BERT model fine-tuned on the original (untranslated) English train split on the test sets obtained by back-translating from each target language.
 
\subsection{Training Details and Hyperparameters}
\label{subsec:training_details}
We implement our experiments using the Hugging Face (HF) Transformers library~\cite{wolf_2020_transformers}, Hydra~\cite{yadan_2019_hydra} and PyTorch~\cite{paszke_2019_pytorch}.\footnote{We make our code publicly available at \url{https://github.com/DFKI-NLP/MultiTACRED} for better reproducibility.} Due to the availability of pretrained models for many languages and to keep things simple, we use BERT as the base PLM~\cite{devlin-etal-2019-bert}. 

We follow~\citet{baldini_soares_matching_2019} and enclose the subject and object entity mentions with special token pairs, modifying the input to become ``\texttt{[HEAD\_START] subject [HEAD\_END]} \ldots \texttt{[TAIL\_START] object [TAIL\_END]}''. In addition, we append the entity types of subject and object to the input text as special tokens, after a separator token: ``\ldots \texttt{[SEP] [HEAD=\textit{type}] [SEP] [TAIL=\textit{type}]}'', where \texttt{\textit{type}} is the entity type of the respective argument.
We use the final hidden state representation of the  \texttt{[CLS]} token as the fixed length representation of the input sequence that is fed into the classification layer. 

We train with batch size of 8 for 5 epochs, and optimize for cross-entropy. The maximum sequence length is 128 for all models. We use AdamW with a scenario-specific learning rate, no warmup, $\beta_1$ = 0.9, $\beta_2 = 0.999$, $\epsilon = 1e-8$, and linear decay of the learning rate. Other hyperparameter values, as well as scenario-specific learning rates and HF model identifiers for the pretrained BERT models, are listed in Appendix~\ref{sec:app_training_details}.

We use micro-F1 as the evaluation metric, and report the median result of 5 runs with different, fixed random seeds. For all experiments, we use the revised version of \textsl{TACRED}  presented by~\citet{alt-etal-2020-tacred}, which fixes a large portion of the \textit{dev} and \textit{test} labels.\footnote{Since both \citet{alt-etal-2020-tacred} and \cite{stoica_2021_retacred} provide fixes as patch files to the original dataset, it is trivial to repeat our experiments using the original or the \textsl{Re-TACRED} version of the data.} We report scores on the test set in the respective target language, denoted as $test_L$. Due to the automatic translation and validation, training and test sets differ slightly across languages, and absolute scores are thus not directly comparable across languages. We therefore also report scores on the intersection test set of instances available in all languages ($test_{\cap}$). This test set contains 11,874 instances, i.e.\ 76.6\% of the original test set (see also Table~\ref{tab:app_multitacred_dataset_stats}).

\section{Results and Discussion}
\label{sec:results_discussion}
We first present some insights into translation quality, and then discuss the performance of models for the different training scenarios.

\subsection{Translation Quality}
\label{subsec:results_translation_quality}
\textbf{Automatic validation} As described in Section~\ref{subsec:translation}, we validate the target language translation by checking whether the entity mention tag markup was correctly transferred. On average, 2.3\% of the instances were considered invalid after translation. By far the largest numbers of such errors occurred when translating to Japanese (9.6\% of translated instances), followed by Chinese (4.5\%) and Spanish (3.8\%). Table~\ref{tab:app_multitacred_dataset_stats} in Appendix~\ref{sec:app_translation_details} gives more details, and shows the number of valid translations for each language, per split and also for the back-translation of the test split. Back-translation incurred only half as many additional errors as compared to the initial translation of the test split into the target language, presumably due to the fact that `hard' examples had already been filtered out during the first translation step.

The validation basically detects two types of alignment errors - missing and additional alignments. An alignment may be missing in the case of pro-drop languages, where the argument is not realized in the translation (e.g.\ Spanish, Chinese), or in compound noun constructions in translations (e.g.\ in German). In other cases, the aligner produces multiple, disjoint spans for one of the arguments, e.g.\ in the case of coordinated conjunctions or compound constructions with different word order in the target language (e.g.\ in Spanish, French, Russian). Table~\ref{tab:app_translation_errors} in Appendix~\ref{sec:app_translation_errors} lists more examples for the most frequent error categories we observed.

\begin{table}[t!]
\centering
\begin{tabular}{lrr}
\toprule
Language & Q1 (yes) & Q2 (yes) \\
\midrule
ar & 85\% & 92\% \\
de & 100\% & 91\% \\
es & 78\%& 91\% \\
fi & 82\%& 81\%\\
fr & 92\%& 93\%\\
hi & 89\% & 67\% \\
hu & 89\% & 48\% \\
ja & 74\% & 89\% \\
pl & 73\% & 93\% \\
ru & 98\% & 89\% \\
tr & 99\% & 90\% \\
zh & 91\% & 80\% \\\midrule
Avg & 87.5\% & 83.7\% \\
\bottomrule
\end{tabular}
\caption{Translation quality, as judged by native speakers. (Q1) Does the translated text meaningfully express the semantic relation of the English original, regardless of minor translation errors? (Q2) Is the overall translation linguistically acceptable for a native speaker?}
\label{tab:translation_quality}
\end{table}

\noindent\textbf{Manual Validation} Table~\ref{tab:translation_quality} shows the results of the manual analysis of translations. 
With regards to Q1, on average 87.5\% of the translations are considered to meaningfully express the relation, i.e.\ as in the original text. Overall translation quality is judged to be good for 83.7\% of the sampled instances on average across languages. The most frequent error types noted by the annotators are again alignment errors, such as aligning a random (neighboring) token from the sentence with an English pronoun argument in pronoun-dropping languages (e.g.\ Polish, Chinese), and non-matching spans (inclusion/exclusion of tokens in the aligned span). Similar errors have also been observed in a recent study by~\citet{chen-etal-2022-frustratingly}. In highly inflecting languages such as Finnish or Turkish, the aligned entity often changes morphologically (e.g.\ possessive/case suffixes).\footnote{Inflection and compounding both ideally could be solved by introducing alignment/argument span boundaries at the morpheme level, but this in turn may raise issues with e.g.\ PLM tokenization and entity masking.} Other typical errors are uncommon/wrong word choices, (e.g.\ due to missing or wrongly interpreted sentence context), and the omission of parts of the original sentence. Less frequent errors include atypical input which was not translated correctly (e.g.\ sentences consisting of a list of sports results), and non-English source text (approx.\ 1\% of the data, see also~\citet{stoica_2021_retacred}). Table~\ref{tab:app_translation_errors} also lists  examples for these error categories.

\subsection{Model Performance}
\label{subsec:results_model_performance}

\begin{table*}[ht!]
\centering
\setlength{\tabcolsep}{5.5pt}
\begin{tabular}{c|r|rrrrrrrrrrrr}
\toprule
Test set & en & ar & de & es & fi & fr & hi & hu & ja & pl & ru & tr & zh \\\midrule
$test_L$ & 77.1 & 74.2 & 74.1 & 75.7 & 76.4 & 75.0 & 65.1 & 71.8 & 71.8 & 73.7 & 73.7 & 74.2 & 75.4 \\
$test_{\cap}$ & 77.5 & 74.5 & 74.6 & 76.1 & 76.6 & 75.4 & 65.9 & 72.4 & 72.5 & 74.3 & 74.8 & 74.5 & 75.3 \\
\bottomrule
\end{tabular}
\caption{Micro-F1 scores on the TACREV dataset for the monolingual setting. The table shows the median micro-F1 score across 5 runs, on the \textit{test} split of the target language ($test_L$), and on the intersection of test instances available in all languages ($test_{\cap}$). %
}
\label{tab:monolingual_results}
\end{table*}

\textbf{Monolingual}
Table~\ref{tab:monolingual_results} shows the results for the monolingual setting. The English BERT model achieves a reference median micro-F1 score of 77.1, which is in line with similar results for fine-tuned PLMs.~\cite{alt-etal-2020-tacred,Chen2022KnowPromptKP,zhou-chen-2022-improved} Micro-F1 scores for the other languages range from 71.8 (Hungarian) to 76.4 (Finnish), with the notable exception of Hindi, where the fine-tuned BERT model only achieves a micro-F1 score of 65.1\footnote{See also Appendix~\ref{sec:app_training_details} for an additional discussion of Hindi performance issues}. As discussed in Section~\ref{subsec:training_details}, results are not directly comparable across languages. However, the results in Table~\ref{tab:monolingual_results} show that language-specific models perform reasonably well for many of the evaluated languages.\footnote{However, as various researchers have pointed out, model performance may be over-estimated, since the models may be affected by ``translationese''~\cite{riley-etal-2020-translationese,graham-etal-2020-statistical}.} Their lower performance may be due to several reasons: translation errors, smaller train and test splits because of the automatic validation step, the quality of the pre-trained BERT model, as well as language-specific model errors. 

Results on the intersection test set $test_{\cap}$ are slightly higher on average, as compared to $test_L$. Relative differences to English, and the overall `ranking' of language-specific results, remain approximately the same. This reaffirms the performance differences between languages observed on $test_L$. It also suggests that the intersection test set contains fewer challenging instances. %
For Hindi, these results, in combination with the low manual evaluation score of 67\% correct translations, suggest that the translation quality is the main reason for the performance loss.

We conclude that for the monolingual scenario, machine translation is a viable strategy to generate supervised data for relation extraction for most of the evaluated languages. Fine-tuning a language-specific PLM on the translated data yields reasonable results that are not much lower than those of the English model for many tested languages.

\noindent\textbf{Cross-lingual}
\begin{table*}[ht!]
\centering
\setlength{\tabcolsep}{5.8pt}
\footnotesize
\begin{tabular}{p{1.2cm}c|r|rrrrrrrrrrrr}
\toprule
Test set / Wikisize & Metric & en & ar & de & es & fi & fr & hi & hu & ja & pl & ru & tr & zh \\\midrule
& P & 76.7 & 72.1 & 75.2 & 74.0 & 76.7 & 74.3 & 76.1 & 76.5 & 78.6 & 76.9 & 70.6 & 73.6 & 73.2 \\
$test_L$ & R & 77.5 & 60.3 & 74.0 & 73.9 & 64.9 & 73.9 & 53.0 & 59.7 & 51.3 & 70.0 & 74.6 & 57.4 & 70.0\\
& F1 & 77.1 & 65.7 & 74.6 & 73.9 & 70.3 & 74.1 & 62.5 & 67.1 & 62.1 & 73.3 & 72.6 & 64.5 & 71.6 \\\midrule
& P & 76.5 & 73.2 & 75.5 & 74.8 & 78.3 & 75.0 & 76.5 & 76.6 & 79.2 & 77.1 & 70.6 & 73.4 & 73.8 \\
$test_{\cap}$ & R & 78.3 & 61.6 & 74.3 & 75.3 & 65.1 & 74.3 & 54.3 & 60.7 & 50.8 & 71.1 & 75.3 & 58.1 & 69.9 \\
& F1 & 77.4 & 66.9 & 74.9 & 75.0 & 71.1 & 74.6 & 63.5 & 67.7 & 61.9 & 74.0 & 72.9 & 64.9 & 71.8 \\\midrule
WikiSize & $\log_2$(MB) & 14 & 10 & 12 & 12 & 9 & 12 & 7 & 10 & 11 & 11 & 12 & 9 & 11 \\
\bottomrule
\end{tabular}
\caption{Micro-Precision, Recall and F1 scores on the TACREV dataset for the cross-lingual setting. The table shows the median scores across 5 runs, on the translated \textit{test} split of the target language ($test_L$) and on the intersection of test instances available in all languages ($test_{\cap}$), when training mBERT on the English \textit{train} split. For reference, the table also shows the size of mBERT's training data in a given language (Wikisize, as $log_2$(MegaBytes), taken from~\citet{wu-dredze-2020-languages}). Languages with less pretraining data in mBERT suffer a larger performance loss.}
\label{tab:crosslingual_results}
\end{table*}
In the cross-lingual setting, micro-F1 scores are lower than in the monolingual setting for many languages (see Table~\ref{tab:crosslingual_results}). The micro-F1 scores for languages well-represented in mBERT's pretraining data (e.g., English, German, Chinese) are close to their monolingual counterparts, whereas for languages like Arabic, Hungarian, Japanese, or Turkish, we observe a  loss of 4.7 to 9.7 F1 points. This is mainly due to a much lower recall, for example, the median recall for Japanese is only 51.3. The micro-F1 scores are highly correlated with the pretraining data size of each language in mBERT: The Spearman rank correlation coefficient of micro-F1 $L_T$ scores  with the WikiSize reported in~\citet{wu-dredze-2020-languages} is $r_s=0.82$ , the Pearson correlation coefficient is $r_p=0.78$ . Hence, languages which are less well represented in mBERT's pretraining data exhibit worse relation extraction performance, as they don't benefit as much from the pretraining. 

Precision, Recall and F1 on the intersection test set $test_{\cap}$ are again slightly better on average than the scores on $test_L$. For Hindi, our results reaffirm the observations made by~\citet{nag-etal-2021-data-bootstrap} for cross-lingual training using only English training data. Our results for RE also confirm prior work on the effectiveness of cross-lingual transfer learning for other tasks (e.g.,~\citet{conneau-etal-2020-unsupervised,pmlr-v119-hu20b}. While results are lower than in the monolingual setting, they are still very reasonable for well-resourced languages such as German or Spanish, with the benefit of incurring no translation at all for training. However, for languages that are less well-represented in mBERT, using a language-specific PLM in combination with in-language training data produces far better results.

\noindent\textbf{Mixed/Multilingual}
\begin{table*}[ht!]
\centering
\footnotesize
\begin{tabular}{c|rrrrrrrrrrrr|r}
\toprule
In-lang data (\%) & ar & de & es & fi & fr & hi & hu & ja & pl & ru & tr & zh &$\overline{\Delta}$\\\midrule
- & 65.7 & 74.6 & 73.9 & 70.3 & 74.1 & 62.5 & 67.1 & 62.1 & 73.3 & 72.6 & 64.5 & 71.6 & - \\\midrule
5 & 68.7 & 74.9 & 73.9 & 70.5 & 74.3 & 67.0 & 68.8 & 69.2 & 72.2 & 73.0 & 67.8 & 72.5 & +1.7 \\
10 & 69.0 & 74.5 & 73.7 & 70.4 & 73.6 & 68.0 & 68.9 & 70.6 & 72.0 & 72.7 & 68.9 & 73.0 & +1.9 \\
20 & 71.0 & 74.4 & 74.5 & 72.2 & 73.9 & 69.9 & 70.2 & 71.7 & 73.3 & 73.3 & 69.2 & 73.0 & +2.9 \\
30 & 71.4 & 74.8 & 74.8 & 72.3 & 74.2 & 70.1 & 71.0 & 72.3 & 72.9 & 73.2 & 70.1 & 73.7 & +3.2 \\
40 & 71.2 & 74.3 & 74.5 & 72.1 & 73.9 & 70.4 & 70.8 & 71.6 & 73.0 & 73.1 & 70.3 & 74.0 & +3.1 \\
50 & 71.2 & 74.7 & 74.4 & 73.0 & 74.4 & 71.8 & 70.9 & 72.6 & 73.1 & 73.3 & 70.3 & 74.8 & +3.5 \\
100 & 73.5 & 75.8 & 75.9 & 73.5 & 75.6 & 72.4 & 72.4 & 73.3 & 74.3 & 75.6 & 71.6 & 75.4 & +4.7\\

\bottomrule
\end{tabular}
\caption{Micro-F1 scores on the TACREV dataset for the mixed/multilingual setting. The table shows the median micro-F1 score across 5 runs, on the translated \textit{test} split of the target language, when training mBERT on the full English \textit{train} split and various portions, from 5\% to 100\%, of the translated target language \textit{train} split. The last column shows the mean improvement across languages, compared to the cross-lingual baseline. Micro-F1 scores improve when adding in-language training data for languages not well represented in mBERT, while other languages mainly benefit when using all of the English and in-language data, i.e.\ essentially doubling the amount of training data (last row).}
\label{tab:multilanguage_results} 
\end{table*}
Table~\ref{tab:multilanguage_results} shows the results obtained when training on both English and varying amounts of target language data. We can observe a considerable increase of mBERT's performance for languages that are not well represented in mBERT's pretraining data, such as e.g.\ Hungarian. These languages benefit especially from adding in-language training data, in some cases even surpassing the performance of their respective monolingual model. For example, mBERT trained on the union of the English and the complete Japanese \textit{train} splits achieves a micro-F1 score of 73.3, 11.2 points better than the cross-lingual score of 62.1 and 1.5 points better than the 71.8 obtained by the monolingual model on the same test data. Languages like German, Spanish, and French don't really benefit from adding small amounts of in-language training data in our evaluation, but show some improvements when adding 100\% of the target language training data (last row), i.e.\ when essentially doubling the size of the training data. Other languages, like Finnish or Turkish, show improvements over the cross-lingual baseline, but don't reach the performance of their monolingual counterpart. 

Our results confirm observations made by ~\citet{nag-etal-2021-data-bootstrap}, who also find improvements when training on a mixture of gold source language data and projected silver target language data. For the related task of event extraction, \citet{yarmohammadi-etal-2021-everything} also observe that the combination of data projection via machine translation and multilingual PLMs can lead to better performance than any one cross-lingual strategy on its own.

\noindent\textbf{Back-translation}
Finally, Table~\ref{tab:backtranslation} shows the performance of the English model evaluated on the back-translated test splits of all target languages. Micro-F1 scores range from 69.6 to 76.1, and are somewhat lower than the score of 77.1 achieved by the same model on the original test set. For languages like German, Spanish, and French, scores are very close to the original, while for Arabic and Hungarian, we observe a loss of approximately 7 percentage points. These differences may be due to the different quality of the MT systems per language pair, but can also indicate that the model cannot always handle the linguistic variance introduced by the back-translation.

\begin{table*}[ht!]
\centering
\begin{tabular}{c|rrrrrrrrrrrr}

\toprule
 Language & ar & de & es & fi & fr & hi & hu & ja & pl & ru & tr & zh \\\midrule
F1  & 69.6 & 76.1 & 75.8 & 73.6 & 75.9 & 73.3 & 70.0 & 72.2 & 74.7 & 74.0 & 72.1 & 74.8 \\
\bottomrule
\end{tabular}
\caption{Median micro-F1 scores across 5 runs of the English BERT model evaluated on the back-translated \textit{test} splits of all languages. Compared to the micro-F1 score of 77.1 on the untranslated English test set, back-translation results are somewhat lower, due to MT system quality and the linguistic variance introduced by the back-translation.}
\label{tab:backtranslation}
\end{table*}

\section{Related Work}
\label{sec:related_work}
\noindent\textbf{Multilingual RE Datasets}
Prior work has primarily focused on the creation of distantly supervised datasets. Dis-Rex~\cite{bhartiya-etal-2021-disrex} and RelX-Distant~\cite{koksal-ozgur-2020-relx} are large, Wikipedia-based datasets, but cover only 4 resp.\ 5 European languages. SMiLER~\cite{seganti-etal-2021-multilingual} covers 14 European languages, but is very imbalanced, both in terms of relation coverage in the different languages and training data per language~\cite{chen-etal-2022-multilingual}.

Manually supervised datasets include BizRel~\cite{khaldi-etal-2022-hows}, consisting of 25.5K sentences labeled with 5 business-oriented relation types, in French, English, Spanish and Chinese, and the IndoRE dataset of 32.6K sentences covering 51 Wikidata relations, in Bengali, Hindi, Telugu and English~\cite{nag-etal-2021-data-bootstrap}. The IndoRE dataset uses MT to transfer manually labeled examples from English to the three other languages, but implements a heuristic to project entity annotations, without any verification step. Other datasets are very small: The RelX dataset contains a manually translated parallel test set of 502 sentences~\cite{koksal-ozgur-2020-relx}.  \citet{arviv-etal-2021-relation} create a small parallel RE dataset of 533 sentences by sampling from \textsl{TACRED} and translating into Russian and Korean.
For the related task of event extraction, datasets worth mentioning are the multilingual ACE 2005 dataset~\cite{walker-etal-2006-ace}, the TAC multilingual event extraction dataset~\cite{ellis-etal-2016-overview}, and the work of \citet{yarmohammadi-etal-2021-everything}.

\noindent\textbf{Machine Translation for Cross-lingual Learning}
MT is a popular approach to address the lack of data in cross-lingual learning~\cite{pmlr-v119-hu20b,nag-etal-2021-data-bootstrap}. There are two basic options - translating target language data to a well-resourced source language at inference time and applying a model trained in the source language~\cite{Asai2018MultilingualER,cui-etal-2019-cross,pmlr-v119-hu20b}, or translating source language training data to the target language, while also projecting any annotations required for training, and then training a model in the target language~\cite{khalil-etal-2019-cross,yarmohammadi-etal-2021-everything,kolluru-etal-2022-alignment}. Both approaches depend on the quality of the MT system, with translated data potentially suffering from translation or alignment errors~\cite{aminian-etal-2017-transferring,ozaki-etal-2021-project,yarmohammadi-etal-2021-everything}. With very few exceptions, using MT for multilingual RE remains underexplored~\cite{faruqui-kumar-2015-multilingual,zou-etal-2018-adversarial,nag-etal-2021-data-bootstrap}.

\noindent\textbf{Multilingual RE} Previous work in cross- and multilingual RE has explored a variety of approaches. \citet{kim-etal-2014-crosslingual} proposed cross-lingual annotation projection, while \citet{faruqui-kumar-2015-multilingual} machine-translate non-English sentences to English, and then project the relation phrase back to the source language for the task of Open RE. \citet{verga-etal-2016-multilingual} use multilingual word embeddings to extract relations from Spanish text without using Spanish training data. In a related approach, \citet{ni-florian-2019-neural} describe an approach for cross-lingual RE that is based on bilingual word embedding mapping. \citet{lin_2017_neural} employ convolutional networks to extract relation embeddings from texts, and propose cross-lingual attention between relation embeddings to model cross-lingual information consistency. \citet{chen-etal-2022-multilingual} introduce a prompt-based model, which requires only the translation of prompt verbalizers. Their approach thus is especially useful in few- and zero-shot scenarios. 

\section{Conclusion}
\label{sec:conclusion}
We introduced a multilingual version of the large-scale \textsl{TACRED} relation extraction dataset, obtained via machine translation and automatic annotation projection. Baseline experiments with in-language as well as cross-lingual transfer learning models showed that MT is a viable strategy to transfer sentence-level RE instances and span-level entity annotations to typologically diverse target languages, with target language RE performance comparable to the English original for many languages.

However, we observe that a variety of errors may affect the translations and annotation alignments, both due to the MT system and  the linguistic features of the target languages (e.g., compounding, high level of synthesis). \textsl{MultiTACRED} can thus serve as a starting point for deeper analyses of annotation projection and RE challenges in these languages. For example, we would like to improve our understanding of RE annotation projection for highly inflectional/synthetic languages, where token-level annotations are an inadequate solution. In addition, constructing original-language test sets to measure the effects of translationese remains an open challenge.

We plan to publish the translated dataset for the research community, depending on LDC requirements for the original \textsl{TACRED} and the underlying TAC corpus. We will also make publicly available the code for the automatic translation, annotation projection, and our experiments.

\section*{Limitations}
A key limitation of this work is the dependence on a machine translation system to get high-quality translations and annotation projections of the dataset. Depending on the availability of language resources and the MT model quality for a given language pair, the translations we use for training and evaluation may be inaccurate, or be affected by translationese, possibly leading to overly optimistic estimates of model performance. In addition, since the annotation projection for relation arguments is completely automatic, any alignment errors of the MT system will yield inaccurate instances. Alignment is at the token-level, rendering it inadequate for e.g.\ compounding or highly inflectional languages. 
Due to the significant resource requirements of constructing adequately-sized test sets, another limitation is the lack of evaluation on original-language test instances. While we manually validate and analyze sample translations in each target language (Section~\ref{subsec:results_translation_quality}) for an initial exploration of MT effects, these efforts should be extended to larger samples or the complete test sets.
Finally, we limited this work to a single dataset, which was constructed with a specific set of target relations (person- and organization-related), from news and web text sources. These text types and the corresponding relation expressions may be well reflected in the training data of current MT systems, and thus easier to translate than relation extraction datasets from other domains (e.g., biomedical), or other text types (e.g., social media). The translated examples also reflect the source language's view of the world, not how the relations would necessarily be formulated in the target language (e.g., use of metaphors, or ignorance of cultural differences).

\section*{Ethics Statement}
We use the data of the original \textsl{TACRED} dataset ``as is''. Our translations thus reflect any biases of the original dataset and its construction process, as well as biases of the MT models (e.g., rendering gender-neutral English nouns to gendered nouns in a given target language). The authors of the original \textsl{TACRED} dataset~\cite{zhang-etal-2017-position} have not stated measures that prevent collecting sensitive text. Therefore, we do not rule out the possible risk of sensitive content in the data. Furthermore, we utilize various BERT-based PLMs in our experiments, which were pretrained on a wide variety of source data. Our models may have inherited biases from these pretraining corpora.

Training jobs were run on a machine with a single NVIDIA RTX6000 GPU with 24 GB RAM. Running time per training/evaluation is approximately 1.5 hours for the monolingual and cross-lingual models, and up to 2 hours for the mixed/multilingual models that are trained on English and target language data.

\section*{Acknowledgements}
We would like to thank David Harbecke, Aleksandra Gabryszak, Nils Feldhus and the anonymous reviewers for their valuable comments and feedback on the paper. We are also very grateful to all the helpful annotators who evaluated the translations: Ammer Ayach, Yuxuan Chen, Nicolas Delinte, Aleksandra Gabryszak, Maria Gonzalez Garcia, Elif Kara, Tomohiro Nishiyama, Akseli Reunamo, Kinga Schumacher, Akash Sinha, and Tatjana Zeen. Finally, we'd like to thank Gabriel Kressin and Phuc Tran Truong for their help with the code base and running the translations and experiments. This work has been supported by the German Federal Ministry for Economic Affairs and Climate Action as part of the project PLASS (01MD19003E), and by the German Federal Ministry of Education and Research as part of the projects CORA4NLP (01IW20010) and Text2Tech (01IS22017B).

\bibliography{main}
\bibliographystyle{acl_natbib}
\appendix

\section{Translation Details}
\label{sec:app_translation_details}
We use the following parameter settings for DeepL API calls:
\textit{split\_sentences:1, tag\_handling:xml, outline\_detection:0}. For Google, we use \textit{format\_:html, model:nmt}.

Table~\ref{tab:app_multitacred_dataset_stats} shows the number of syntactically valid and invalid translations for each language and split, as well as for the back-translation of the test split.

For tokenization, we use Spacy 3.2\footnote{\url{https://spacy.io}} with standard (non-neural) models for \textit{de, es, fr, fi, ja, pl, ru, zh}, and TranKIT 1.1.0\footnote{\url{https://github.com/nlp-uoregon/trankit}} for \textit{ar, hi, hu, tr}.

The translation costs per language amount to approximately 460 Euro, for a total character count of 22.9 million characters to be translated (source sentences including entity markup tags), at a price of 20 Euro per 1 million characters at the time of writing. Compared to an estimated annotation cost of approximately 10K USD, translation costs amount to less than 5\% of the cost of fully annotating a similar-sized dataset in a new language.\footnote{\citet{stoica_2021_retacred} pay 0.15 USD per HIT of 5 sentences in \textsl{TACRED}. With an average of 3 crowd workers per HIT and a total of 106,264 examples in \textsl{TACRED}, this amounts to approximately 9,564 USD. \citet{angeli-etal-2014-combining} report a cost of 3,156 USD for annotating 23,725 examples, which would correspond to a cost of 14,135 USD for the whole \textsl{TACRED} dataset.}

\begin{table*}[ht!]
\centering
\footnotesize
\begin{tabular}{p{2cm}rrrrrrrr}
\toprule
& Train & Train Err & Dev & Dev Err & Test & Test Err & BT Test & BT Test Err \\
Language (Translation Engine)  &  &  &  &  &  &  &  &  \\
\midrule
en (-) & 68,124 & - & 22,631 & - & 15,509 & - & - & - \\
ar (G) & 67,736 & 388 & 22,502 & 129 & 15,425 & 84 & 15,425 & 0 \\
de (D) & 67,253 & 871 & 22,343 & 288 & 15,282 & 227 & 15,079 & 203 \\
es (D) & 65,247 & 2,877 & 21,697 & 934 & 14,908 & 601 & 14,688 & 220 \\
fi (D) & 66,751 & 1,373 & 22,268 & 363 & 15,083 & 426 & 14,462 & 621 \\
fr (D) & 66,856 & 1,268 & 22,298 & 333 & 15,237 & 272 & 15,088 & 149 \\
hi (G) & 67,751 & 373 & 22,511 & 120 & 15,440 & 69 & 15,440 & 0 \\
hu (G) & 67,766 & 358 & 22,519 & 112 & 15,436 & 73 & 15,436 & 0 \\
ja (D) & 61,571 & 6,553 & 20,290 & 2,341 & 13,701 & 1,808 & 12,913 & 805 \\
pl (G) & 68,124 & 0 & 22,631 & 0 & 15,509 & 0 & 15,509 & 0 \\
ru (D) & 66,413 & 1,711 & 21,998 & 633 & 14,995 & 514 & 14,703 & 292 \\
tr (G) & 67,749 & 375 & 22,510 & 121 & 15,429 & 80 & 15,429 & 0 \\
zh (D) & 65,260 & 2,864 & 21,538 & 1,093 & 14,694 & 815 & 14,021 & 681\\ 
\midrule
$\cap_{all}$ & 54,251 & - & 17,809 & - & 11,874 & - & 9,944 & - \\
\bottomrule
\end{tabular}
\caption{\textsl{MultiTACRED} instances per language and split, and for the back-translation (BT) of the test split. The \emph{`en'} row shows the statistics of the original \textsl{TACRED}. (G) and (D) refer to Google and DeepL, respectively. The error columns list the number of instances discarded after translation due to missing / erroneous entity tag markup. On average, 2.3\% of the instances were discarded due to invalid entity markup after translation. The last row shows the intersection of valid instances available in all languages.}
\label{tab:app_multitacred_dataset_stats}
\end{table*}

\section{Human Translation Analysis}
\label{sec:app_translation_analysis}
\begin{figure*}[ht!]
\centering
\includegraphics[width=0.6\textwidth,keepaspectratio]{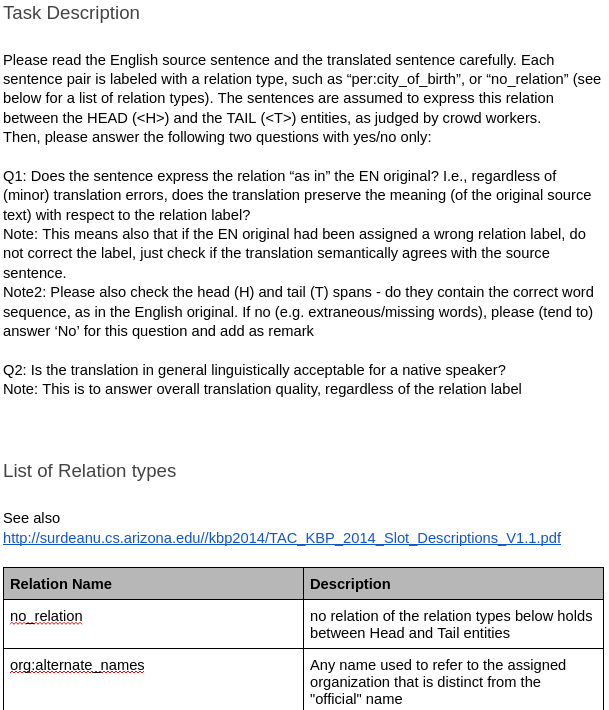}
\caption{Task description given to human judges for translation quality analysis.}
\label{fig:app_annotator_instructions}
\end{figure*}

For the manual analysis of translated TACRED instances, we recruited a single native speaker for each language among the members of our lab and associated partners. Annotators were not paid for the task, but performed it as part of their work at the lab. All annotators are either Master's degree or PhD students, with a background in Linguistics, Computer Science, or a related field. The full instructions given to annotators, after a personal introduction to the task, are shown in Figure~\ref{fig:app_annotator_instructions}.

\section{Additional Training Details}
\label{sec:app_training_details}
All pre-trained models evaluated in this study are used as they are available from HuggingFace's model hub, without any modifications. Our implementation uses HF's \textit{BertForSequenceClassification} implementation with default settings for dropout, positional embeddings, etc. Licenses for the pretrained BERT models are listed in Table~\ref{tab:app_models_hyperparams}, if specified in the repository. The Transformers library is available under the Apache 2.0 license, Hydra under the MIT license, and PyTorch uses a modified BSD license.

For Hungarian, we use \textit{bert-base-multilingual-cased}, since there is no pretrained Hungarian BERT model available on the hub. For Hindi, we tried several models by l3cube-pune, neuralspace-reverie, google and ai4bharat, but all of these produced far worse results than the ones reported here for \textit{l3cube-pune/hindi-bert-scratch}. Interestingly, using \textit{bert-base-multilingual-cased} instead of \textit{l3cube-pune/hindi-bert-scratch} as the base PLM produced far better results for Hindi in the monolingual setting, at 71.1 micro-F1. 

\begin{table*}[ht!]
\centering
\footnotesize
\begin{tabular}{llll}
\toprule
Language/Scenario & HuggingFace Model name & LR & License \\
\midrule
ar & aubmindlab/bert-base-arabertv02 & 1e-5 & N/A \\
de & bert-base-german-cased & 3e-5 & MIT \\
en & bert-base-uncased & 3e-5 & Apache 2.0 \\
es & dccuchile/bert-base-spanish-wwm-cased & 1e-5 & (CC BY 4.0) \\
fi & TurkuNLP/bert-base-finnish-cased-v1 & 7e-6 & N/A  \\
fr & flaubert/flaubert\_base\_cased & 1e-5 &  MIT \\
hi & l3cube-pune/hindi-bert-scratch & 7e-6 & CC BY 4.0 \\
hu & bert-base-multilingual-cased & 1e-5 & Apache 2.0 \\
ja & cl-tohoku/bert-base-japanese-whole-word-masking & 3e-5 & CC BY 4.0 \\
pl & dkleczek/bert-base-polish-cased-v1 & 7e-6 & N/A \\
ru & sberbank-ai/ruBert-base & 3e-5 & Apache 2.0 \\
tr & dbmdz/bert-base-turkish-cased & 1e-5 & MIT \\
zh & bert-base-chinese & 1e-5 & N/A \\
\midrule
Cross-lingual mBERT & bert-base-multilingual-cased & 1e-5 & Apache 2.0  \\
\midrule
Multilingual mBERT & bert-base-multilingual-cased & 1e-5 & Apache 2.0  \\
\bottomrule
\end{tabular}
\caption{Best learning rate and model identifiers per language for the monolingual settings, and for the cross- and multilingual scenarios. The table also lists the model license, if it was available.}
\label{tab:app_models_hyperparams}
\end{table*}

We experimented with learning rates in $[3e-6, 7e-6, 1e-5, 3-e5, 5e-5]$. We used micro-F1 on the \textit{dev} set as the criterion for hyperparameter selection. Table~\ref{tab:app_models_hyperparams} lists the best learning rates per language and scenario. We use a fixed set of random seeds \{1337, 2674, 4011, 5348, 6685\} for training across the 5 runs.

\section{Translation Error Examples}
\label{sec:app_translation_errors}
Table~\ref{tab:app_translation_errors} lists common error types we identified in the translations of \textsl{TACRED} instances.
\begin{table*}[ht!]
\centering
\scriptsize
\begin{tabular}{p{1.2cm}p{4.7cm}cp{4.7cm}p{2.7cm}}

\toprule
\textbf{Error Type} & \textbf{Source} & \textbf{Lang.} & \textbf{Translation} & \textbf{Comment} \\\midrule
Alignment - Missing & <H>He</H> also presided over the country 's <T>Constitutional Council</T> [\ldots] & es & También presidió el <T>Consejo Constitucional</T> del país [\ldots] & Head not marked due to dropped pronoun \\
Alignment - Definite Article & <T>JetBlue Airways Corp</T> spokesman <H>Bryan Baldwin</H> said [\ldots] & es & <H>El</H> portavoz de<T>JetBlue Airways Corp</T> <H>, Bryan Baldwin</H>, dijo [\ldots] & `El' is marked as additional head span\\
Alignment - Split span & New <T>York-based Human Rights Watch</T> ( HRW ) ,  [\ldots] snubbed an invitation to testify [\ldots] & es 
 &	 <T>Human Rights Watch</T> (HRW), con sede en Nueva <T>York</T>, [\ldots] rechazaron una invitación para testificar [\ldots]
 & Translation of `York-based' syntactically different, leading to split span \\
Alignment - Split Compound &  [\ldots] Russian <T>Foreign Ministry</T> spokesman Andrei Nesterenko said on Thursday , <H>RIA Novosti</H> reported. & fr &  [\ldots] a déclaré jeudi le porte-parole du <T>ministère</T> russe <T>des affaires étrangères</T>, Andrei Nesterenko, selon <H>RIA Novosti</H>. & French word order for adjectives leads to split span of compound `Foreign Ministry'  \\
Alignment - Compound &  [\ldots] Seethapathy Chander , Deputy Director General with <T>ADB</T> 's <H>Private Sector Department</H>.  & de  &  [\ldots] Seethapathy Chander, stellvertretender Generaldirektor der <H>ADB-Abteilung für den Privatsektor</H>. & German translation uses a compound noun combining head and `department' \\
Alignment - Missing & <H>She</H> was vibrant , she loved life and <T>she</T> always had a kind word for everyone. & de &	<H>Sie</H> war lebhaft, sie liebte das Leben und hatte immer ein freundliches Wort für jeden. & Multiple occurrences of same pronoun seem to confuse aligner \\
Alignment - Coordination & <H>Christopher Bentley</H> , a spokesman for Citizenship and <T>Immigration Services</T> [\ldots] & es  & <H>Christopher Bentley</H>, un portavoz de <T>los Servicios de</T> Ciudadanía e <T>Inmigración</T> [\ldots]
 & Coordinated conjuction in Spanish leads to split span \\\midrule
Alignment - Wrong & She said when <H>she</H> got pregnant in <T>2008</T>  [\ldots] & pl & Powiedziała, że kiedy w <T>2008</T> r. <H>zaszła</H> w ciążę [\ldots] & `got' marked instead of dropped pronoun `she' \\
Alignment - Extended & <T>Alaskans</T> last chose a Democrat for the presidency in 1964 , when they backed Lyndon B. Johnson by a 2-1 margin over <H>Barry Goldwater</H> . & zh  & \begin{CJK*}{UTF8}{gbsn} <T>阿拉斯加人上</T>一次选择民主党人担任总统是在1964年，当时他们以2比1的优势支持林登-B-约翰逊，而不是<H>巴里-戈德华特</H>。 \end{CJK*} & `last' is included in tail span \\
Alignment - Partial & In August , <H>Baldino</H> [\ldots] had taken a leave of absence from his posts as Cephalon 's chairman and <T>chief executive</T> . & pl & W sierpniu <H>Baldino</H> [\ldots] wziął urlop od pełnienia funkcji prezesa i <T>dyrektora general</T> nego firmy Cephalon. & `nego' should be part of the tail span and not be split off of the word `generalnego' \\
Alignment - Inflection & Some of the people profiled are <T>ABC</T> president <H>Steve McPherson</H> , [\ldots] & fi & Mukana ovat muun muassa <T>ABC:n</T> pääjohtaja <H>Steve McPherson</H>, [\ldots] & Tail 'ABC:n' includes genitive case marker in Finnish\\ 
Non-English Source & Dari arah Jakarta/Indramayu , <T>sekitar</T> 2 km sebelum Pasar Celancang , tepatnya di sebelah Kantor Kecamatan Suranenggala terdapat Tempat Pelelangan Ikan ( <H>TPI</H> ) . & - & - & Source language is Indonesian, not English \\
Sentence split & <H>Stewart</H> is not saying that a 1987-style stock market crash is on the immediate horizon , and <T>he</T> concedes that `` by many measures , stocks are n't overpriced , even at recent highs . '' & tr & <H>Stewart</H>, 1987 tarzı bir borsa çöküşünün hemen ufukta olduğunu söylemiyor ve <T>o</T> `` birçok önlemle , hisse senetlerinin aşırı fiyatlandırılmadığını bile kabul ediyor . son zirvelerde. '' & `son zirvelerde' erroneously separated by end-of-sentence period \\
Translation incomplete & Outlined in a filing with the <H>Federal Election Commission</H> , <T>Obama</T> 's suggestion is notable because \ldots & de &	Der Vorschlag <T>Obamas</T> ist bemerkenswert, weil \ldots & Translation is missing first part and head span \\
Atypical input &  Browns 5-10 [\ldots] <T>Cowboys</T> 5-10 [\ldots] <H>Jaguars</H> 8-7  [\ldots] Total : 42-93 ( .311 ) Total : 58-74 ( .439 ) Total : 53-81 ( .396 ) & zh & \begin{CJK*}{UTF8}{gbsn} Browns 5-10  [\ldots] <T>Cowboys</T>5-10  [\ldots] <H>Jaguars</H>8-7  [\ldots] Total : 42-93 ( .311 ) 总数 : 58-74 ( .439 ) 总数 : 53-81 ( .396 ) \end{CJK*} & Almost no translation due to atypical input\\
\bottomrule
\end{tabular}
\caption{Common error types of translated \textsl{TACRED} examples. The first half of the table shows alignment errors that can be automatically detected, such as missing or additional aligned spans in the translation. The second half shows error types identified by human judges.}
\label{tab:app_translation_errors}
\end{table*}

\end{document}